\documentclass[conference]{IEEEtran}
\IEEEoverridecommandlockouts
\usepackage{hyperref}
\usepackage{cite}
\usepackage{amsmath,amssymb,amsfonts}
\usepackage{textcomp}
\usepackage{xcolor}
\def\BibTeX{{\rm B\kern-.05em{\sc i\kern-.025em b}\kern-.08em
    T\kern-.1667em\lower.7ex\hbox{E}\kern-.125emX}}

\usepackage{algorithm}
\usepackage[noend]{algpseudocode}
\usepackage{multirow}

\usepackage{graphicx}
\usepackage{subcaption}

\usepackage{tikz}

\usepackage{soul}

\begin{document}

\title{A Bayesian optimization framework for the automatic tuning of MPC-based shared controllers\\
}
\author{\fontsize{10.5}{11} \selectfont Anne van der Horst $^{1*}$, Bas Meere$^{1*}$,  Dinesh Krishnamoorthy$^{1}$,  \\ Saray Bakker$^{1,2}$, Bram van de Vrande$^{3}$, Henry Stoutjesdijk$^{3}$, Marco Alonso$^3$ and Elena Torta$^{1}$
\thanks{$^{*}$Equal contribution.}
\thanks{$^{1}$
        Eindhoven University of Technology, The Netherlands (e-mail:\tt{ \small\href{mailto:a.v.d.horst1@tue.nl}{a.v.d.horst1@tue.nl}})}
\thanks{$^{2}$ Delft University of Technology, The Netherlands}
\thanks{$^{3}$Philips IGT Systems Mechatronics, The Netherlands}%
}

\maketitle

\begin{abstract}
This paper presents a Bayesian optimization framework for the automatic tuning of shared controllers which are defined as a Model Predictive Control (MPC) problem.  The proposed framework includes the design of performance metrics as well as the representation of user inputs for simulation-based optimization.
The framework is applied to the optimization of a shared controller for an Image Guided Therapy robot. VR-based user experiments confirm the increase in performance of the automatically tuned MPC shared controller with respect to a hand-tuned baseline version as well as its generalization ability.
\end{abstract}


\section{Introduction}
\begin{figure}
    \centering\includegraphics[width=0.94\linewidth]{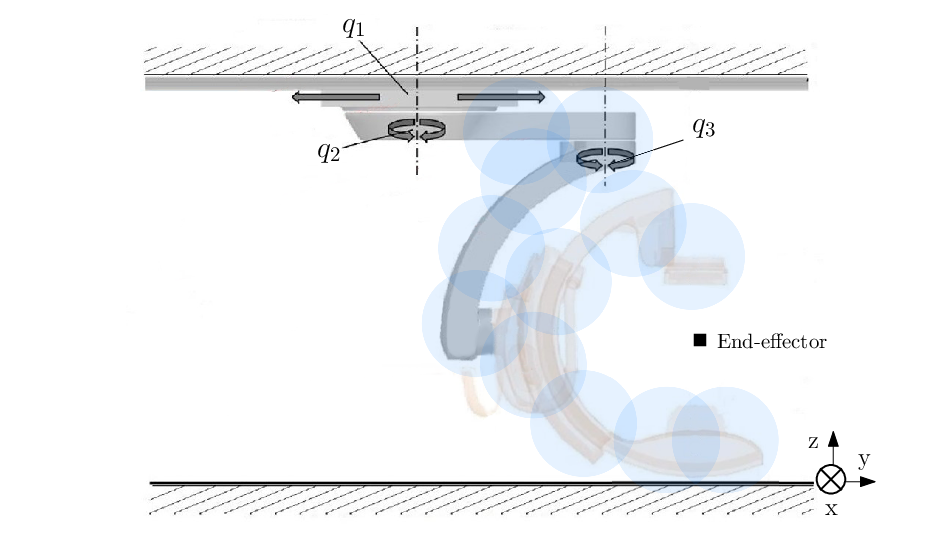}
    \caption{The IGT robot with its prismatic joint $q_1$, two revolute joints $q_2$ and $q_3$ and the end-effector position. The 12 spheres for obstacle avoidance capture the relevant geometry of the robot.}
    \label{fig:drawing_robot}
\vspace{-0.25in}
\end{figure}

Shared control \cite{abbink2018topology} is a paradigm for robot semi-autonomous task execution which requires the graceful blending of users' and robots' own decision-making to achieve a shared task. 
For semi-autonomous navigation, typically users provide desired navigation targets or movement directions and the robot performs obstacle avoidance autonomously while trying to reach the user's goal.
The performance of such a control algorithm often depends on the tuning of its parameters. The tuning process notoriously takes the form of a trial-and-error approach which is time-consuming, error-prone, and can become frustrating.
Bayesian Optimization (BO) \cite{frazier2018bayesian} is a data-based optimization technique that has surged in popularity as a viable way of automatically tuning controller parameters for robotic systems (e.g., \cite{berkenkamp2021bayesian, 7989186, marco2016automatic}). 
Contrary to Reinforcement Learning, BO is a gradient-free method and typically reaches convergence with fewer iterations. Both aspects are important to incorporate direct user feedback in the tuning process of algorithms for human-robot interaction tasks.
\newline
While previous work has addressed the development of BO algorithms for autonomous motion tasks of manipulators \cite{marco2016automatic, 7989186} and drones with emphasis on safety constraints \cite{berkenkamp2021bayesian, brunke2022safe}, little attention has been given to controller optimization for human-robot interaction tasks, such as shared control. 
\newline
The contribution of this paper relates to the definition of a BO framework for the automatic tuning of shared controllers for robot navigation. In addition to the design choices of the algorithm itself,  distinctive challenges relate to the definition of representative user inputs and the identification of the performance metrics for the optimization process. 
\newline
The proposed BO framework is applied to the optimization of an MPC-based shared controller for an  Image Guided Therapy (IGT) robot, shown in Fig.~\ref{fig:drawing_robot}. 
These robots are used to create medical images via X-ray during minimally invasive procedures and are constituted by several links with the terminal link presenting the classical C-shape carrying the X-ray generator and detector. 
In the current state-of-practice,  the (re)positioning of the robot in the operating room requires inputs from the physician through a joystick-like control module. The development of a shared controller is expected to facilitate positioning because the robot can automatically avoid obstacles while following users' commands as closely as possible. 
The motion performance of the automatically-tuned MPC shared controller is compared with a baseline, hand-tuned controller, by means of Virtual Reality (VR) experiments involving users interacting with the system through a joystick-like control module. The comparison is based on metrics that measure the safety and smoothness of the trajectories as well as the efficiency of the controller. Results show an overall performance improvement of the automatically tuned shared controller w.r.t. the baseline as well as a good generalization for user input sequences not used during optimization. 
\newline
\textit{Related Work: } BO has been successfully applied for the automatic tuning of controllers for several robotic tasks. One of the earlier examples is the work of Marco et al. \cite{marco2016automatic} that optimized an LQR controller for an inverted pendulum actuated by a robotic arm. It showed the successful application of BO to a physical robotic system. Subsequent work has focused on safe BO \cite{berkenkamp2021bayesian, brunke2022safe} which aims at enabling a safe exploration strategy of the algorithm's tuning parameters. The latter is particularly relevant for the application of the optimization framework  on physical systems or with direct user's involvement. 
Recent applications of BO to robotic systems also investigated its relation to the selection of rewards for reinforcement learning algorithms (e.g., \cite{jaquier2020bayesian, turchetta2020robust}) as well as the automatic tuning of MPC-based controllers (e.g., \cite{edwards2021automatic, frohlich2022contextual, yeganegi2021robust}).
\newline
Edwards et al. \cite{edwards2021automatic} devised \texttt{AutoMPC}; a software package to jointly identify the underlying system dynamics and automatically tune the parameters of data-driven MPC controllers for robotics applications. 
\texttt{AutoMPC} was validated in an OpenAI Gym environment. 
An example involving the application of BO to the automatic tuning of an MPC controller on a scaled model of a racing car is proposed in \cite{frohlich2022contextual}.
\newline
Semi-autonomous navigation under shared control requires a seamless integration of inputs from a human operator and an autonomous controller. To this end, MPC is an advantageous technique as it can incorporate constraints on a robot's actuation limits as well as predictions of future system's states and user's commands. Several examples of MPC-based shared controllers for semi-autonomous robot navigation are proposed in the literature (e.g., \cite{storms2014blending, zarei2020experimental, chipalkatty2010human}).
In \cite{storms2014blending}, the controller is employed to blend human input and autonomous behavior to enable the autonomous navigation of a mobile robot in a cluttered environment. 
Another example of an MPC-based shared controller is given in the experimental study of Zarei et al. \cite{zarei2020experimental}, where a mobile robot could navigate semi-autonomously by minimizing the deviations from a human input while upholding obstacle avoidance constraints.
Taking one step further, Chipalkatty et al. \cite{chipalkatty2010human} design the control problem such that it includes the system dynamics to minimize deviations from a human input while safeguarding a linear state constraint. 
Since the current paper aims to apply the shared controller to a robot that resembles a mobile manipulator, we designed the controller based on an adaptation of the perceptive MPC architecture for mobile manipulators presented in \cite{pankert2020perceptive}. 


\section{Shared control framework}\label{sec:shared_control}
\begin{figure}
    \centering
    \includegraphics[width = 0.45\textwidth]{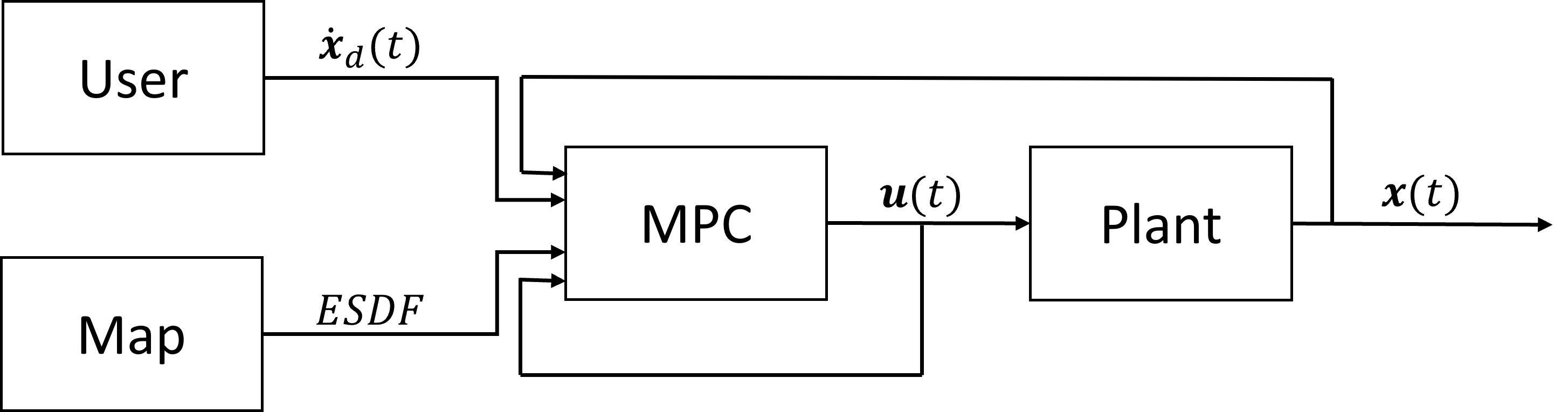}
    \caption{The control architecture of the proposed MPC-based shared controller.}
    \label{fig:control_loop}
    \vspace{-0.25in}
\end{figure}

The  shared controller for robot semi-autonomous navigation is designed to optimize the tradeoff between following user commands as closely as possible while autonomously avoiding obstacles.
The architecture proposed in \cite{pankert2020perceptive} has been adapted for a navigation task under shared control by considering a target velocity provided by the user. The block diagram of the resulting architecture is illustrated in Fig~\ref{fig:control_loop}.
\newline
User input is gathered from a joystick-like control module in the form of a reference velocity vector, i.e., $\dot{\boldsymbol{x}}_{d}$, which represents the velocity the robot is required to track. The reference velocity is applied to the robot's end-effector which, specifically for the IGT robot,  is defined as the midpoint between the X-ray generator and detector (see Fig.~\ref{fig:drawing_robot}).
\newline
A voxel map of the environment is created which is subsequently converted into a Euclidean Signed Distance Field (ESDF) \cite{maurer2003linear} that reports, for every voxel, the distance to the closest obstacle. For computational efficiency, the entire geometry of the robot is approximated using a set of spheres (as depicted in Fig.~\ref{fig:drawing_robot}). The number of spheres for the specific IGT robot, $n_{cs}$,  was $12$, all with a radius $r_{cs} = 0.4$ m. Relying on the gradient of the ESDF, it is possible to derive the preferred direction to move each sphere away from obstacles.
\newline
The control effort, in the form of joints' velocities, is computed by solving a non-linear MPC problem  whose  expression  is given by the following equations:
\begin{align}
    \min\limits_{x,u} \quad & \alpha \sum\limits_{k=0}^{N_{c}-1} l\bigl(\boldsymbol{x}(k),\boldsymbol{u}(k)\bigr)+ (1 - \alpha) \sum\limits_{k=1}^{N_p} \sum\limits_{m=1}^{n_{c s}} B_m\bigl(\boldsymbol{x}\left(k\right)\bigr), \label{eq:stage_costs} \\
    \text{s.t.} \quad & \boldsymbol{x}\left(k+1\right)=\boldsymbol{x}\left(k\right)+T_s f\bigl(\boldsymbol{x}\left(k\right), \boldsymbol{u}\left(k\right)\bigr), \label{eq:discrete_system} \\
    & \boldsymbol{x}\left(0\right)=x_0, \label{eq:init} \\
    & \boldsymbol{h}\bigl(\boldsymbol{x}\left(k\right), \boldsymbol{u}\left(k\right)\bigr) \geq 0. \label{eq:constr}
\end{align}

%
The state vector is defined as $\boldsymbol{x}(k) = \begin{bmatrix} \boldsymbol{x}_{e}(k) & \boldsymbol{q}(k) \end{bmatrix}^{T}$ with $\boldsymbol{x}_{e}(k)$ representing the 2D pose of the end-effector and $\boldsymbol{q}(k)$ representing the vector of the positions of the joints. 
For the IGT robot, $\boldsymbol{q}(k)$ represents the positions of the prismatic joint $q_1$ and the revolute joints $q_2$ and $q_3$ (see Fig.~\ref{fig:drawing_robot}). The control effort, $\boldsymbol{u}(k)$, is chosen as the velocity of the joints.
The stage cost in Eq.~\ref{eq:stage_costs} includes  a term to penalize the deviation of the end-effector's Cartesian velocity from the joystick input provided by the user: 
\begin{equation}
    l\bigl(\boldsymbol{x}(k),\boldsymbol{u}(k)\bigr) = \left\|\boldsymbol {J}_{e}\bigl(\boldsymbol{x}(k)\bigr)\boldsymbol{u}(k) - \dot{\boldsymbol{x}}_{d}(k)\right\|_{\boldsymbol{Q}}^2,
    \label{eq:ref_tracking}
\end{equation}
where $\boldsymbol{J}_{e}$ is the Jacobian matrix of the robot and $\boldsymbol{Q} \in \mathbb{R}^{3 \times 3}$ a positive definite weight matrix. The reference input $\dot{\boldsymbol{x}}_{d}(k)$ is assumed constant over the prediction horizon $N_p$. 
The other term of the stage cost penalizes joint configurations that bring the collision spheres closer to obstacles. 
Its expression is given by:  
\begin{equation}
    B_{m}\bigl(\boldsymbol{x}\left(k\right)\bigr) = \frac{c_{1}}{1+\mathrm{e}^{c_{2}(sd(\boldsymbol{x}(k))-c_{3})}},
    \label{eq:penalty_MPC}
\end{equation}
which is taken as a function of the state vector.
Here, $sd(\boldsymbol{x}(k))$ is the signed distance of the center of each collision sphere to the closest obstacle, and $c_{1}, c_{2}$ and $c_{3}$ are scaling factors. In this work, $c_3 = r_{cs}$.
Individual contributions are summed in Eq.~\ref{eq:stage_costs} over the entire prediction horizon $N_p$.
The blending scalar, a typical element of shared controllers that specifies the priority between tracking user input and avoiding obstacles,  is defined as $\alpha$, which is a scalar between 0 and 1. In this study, $\alpha = 0.5$, to assign equal value to the human input and obstacle avoidance term.
In our formulation, the control horizon $N_c$ can be different from the prediction horizon $N_p$.
Furthermore, Eq.~\ref{eq:discrete_system} discretizes the forward kinematics of the system $\forall k \in N_p$, where $T_{s}$ denotes the discrete sample time using the zero-order hold method \cite{landau2006digital}. The initial conditions are defined in Eq.~\ref{eq:init}, while Eq.~\ref{eq:constr} contains position, velocity, acceleration and jerk constraints on the joints of the robot as well as the end-effector. The constraints are summed $\forall k \in N_p, N_c$, depending on whether the constraint relies on states $\boldsymbol{x}(k)$ or inputs $\boldsymbol{u}(k)$.
\newline
The choice of the MPC parameters for automatic tuning depends on several factors such as the robot's geometry and the task's specifications. Due to scalability issues of BO \cite{letham2020re, frazier2018bayesian}, it is recommended to keep the amount of tuning parameters low. For the validation use-case, a sensitivity analysis with the elementary effect method~\cite{binois2022survey} was performed which resulted in the selection of the following parameters; the prediction and control horizon $N_p$ and $N_c$, the three elements of the $\boldsymbol{Q}$ matrix in Eq.~\ref{eq:ref_tracking} that penalizes differences between reference end-effector velocities and computed velocities, and the two scaling variables of the obstacle avoidance term, i.e., $c_1$ and $c_2$ of Eq.~\ref{eq:penalty_MPC}. To summarize, seven tuning parameters were selected, given by $\boldsymbol{\xi} = [N_p, N_c, Q_{x}, Q_{y}, Q_{\theta},  c_1, c_2]$. The choice of tuning parameters is in line with related works. For example, selecting the individual elements of the matrix $\boldsymbol{Q}$ is quite common practice (e.g., \cite{mehndiratta2018automated, gharib2021multi}). The selection of the control $N_c$ and prediction horizon $N_p$ is also reported in \cite{kapnopoulos2022cooperative}.

\section{Bayesian optimization framework}\label{sec:BO_method}
In this section, we describe the proposed setup of the BO framework for the shared controller. For background information on BO and its theoretical underpinning, the reader is referred to \cite{frazier2018bayesian, shahriari2015taking}.
\newline
We cast the BO problem as a multi-objective constraint optimization problem:
\begin{align}\label{eq:BO}
\begin{aligned}
     \min_{\boldsymbol{\xi}} J_{\xi} = \qquad & {n_{mov}}\sum^{n_{mov}}_{j=1} \sum^{n_{obj}}_{h=1} w_hF_{hj}(\boldsymbol{\xi}), \\
    \text { s.t.} \qquad & N_p \geq N_c, \\
     & n_{succ}(\boldsymbol{\xi}) = n_{mov}, \\
     & sd(\boldsymbol{x}) \geq r_{cs}.
\end{aligned}
\end{align}

%
%
%
The term $\boldsymbol{\xi}$ represents the vector of tunable MPC parameters. 
  we assume the objective function can be evaluated precisely only at the point $ \boldsymbol{\xi}_{i}$ selected for the $i-th$ optimization iteration~\cite{frazier2018bayesian}. The objective function is thus approximated by a statistical surrogate model which commonly takes the form of a  Guassian Process \cite{frazier2018bayesian, shahriari2015taking}. For this work, the kernel of the Gaussian Process is chosen as  a Mat\'ern 5/2 kernel \cite{shahriari2015taking}. The next sampling set, $ \boldsymbol{\xi}_{i}$, is determined by maximizing the acquisition function of which an analytical expression is known. The acquisition function can take different expressions. For the validation use-case,  we opted for the Expected Improvement (EI) \cite{frazier2018bayesian}. 
The value of the objective function, $J_{\xi}$,  at the sampling point $ \boldsymbol{\xi}_{i}$, can be evaluated by means of experiments in which the performance of the robot's motion is quantified with a set of $n_{obj} = 6$ performance metrics (i.e., $F_{hj}(\boldsymbol{\xi}) $, where $h$ denotes the metric index and $j$ the movement index). The proposed analytical formulation of $F_{hj}(\boldsymbol{\xi}) $ is  described in detail in Section~\ref{sec:objective_function}.  The relative importance of the metrics is expressed through the weights $w_h$. 
At every iteration, $n_{mov}$ movements are performed with the MPC controller configured with the current parameter set $ \boldsymbol{\xi}_{i}$. 
At every iteration of the optimization algorithm, the performance metrics are computed for each movement and summed over the total number of movements $n_{mov}$.
For the validation setup reported in the paper, movements were performed in a simulation environment. Executing a movement in simulation requires the formulation of a user input which we elaborate on more extensively in  Section~\ref{sec:user_input}.
The full procedure is reported in Algorithm~\ref{alg:BO}.
\begin{algorithm}
    \caption{BO procedure for shared controller tuning.}
    \label{alg:BO}
    \begin{algorithmic}[1]
        \State \text{Define a set of $n_{\text{mov}}$  movements (see Section~\ref{sec:user_input})}
        \State $\boldsymbol{\xi} \gets \xi_{0}$ \text{(initialize tuning parameters)}
        
        \For{$i=0 \textbf{ to }  n_{\text{max}}$} \text{(optimization iterations)}
            \For{$j=0$ \textbf{ to } $n_{\text{mov}}$} 
                \State \text{Perform a movement,  evaluate metrics}
            
            \EndFor
            \State Compute objective function for iteration $i$ ( Eq.~\ref{eq:BO}) 
            \State Update the surrogate function
            \State Update the acquisition function
            \State Select a new value of $\boldsymbol{\xi}$ by maximizing the acquisition function
        \EndFor
        \State \textbf{return} $\boldsymbol{\xi}$ with the smallest objective function
    \end{algorithmic}
\end{algorithm}
For the validation use-case, the procedure was implemented in Matlab using the statistics and machine learning toolbox. Simulations were performed in Simulink Simscape.

\subsection{Metrics definition}\label{sec:objective_function}
Based on prior work on the evaluation of mobile robot local planners \cite{wen2021mrpb} and pure-motion tasks of mobile robots \cite{calisi2009performance}, we identified $n_{obj} = 6$ metrics by which to evaluate the performance of the MPC-based shared controller. Each metric relates to one of three fundamental aspects for human-robot interaction namely safety, smoothness and efficiency.
\newline
\textit{Safety} is arguably one of the most important aspects since shared control is often deployed in robots that operate in human-populated environments. We identified two metrics to evaluate safety, i.e., obstacle proximity and time spent close to obstacles.
\newline
The obstacle proximity metric, denoted as $d_{ob}$, serves as an indicator of safety as it represents the closest approach of the robot to a potential collision during its movement. The metric is calculated as the minimum distance between any link of the robot and obstacles throughout the entire motion of the robot:
        \begin{align}
        F_1:\; d_{ob}=\min_g \left\{\min_m \left\{d_{g,m}\right\}\right\},\; 1 \leq g \leq N,\; 1 \leq m \leq n_{cs},
        \label{eq:d_ob}
        \end{align}
        where $g$ is the index of the sample, $N$ is the total number of samples in a movement, and $m$ is the index of the link. For the validation use-case, the links of the robots are approximated with the collection of spheres, $n_{cs}$, with the distance measured from the center of each sphere.  
\newline
The second safety metric, $t_{ob}$, quantifies the time that the robot spends within a certain distance from the obstacles during a movement. The metric is calculated as the fraction of time spent with any of the links of the robot closer than a threshold $d_{\text {safe}}$ to obstacles. Its expression is given by:
        \begin{align}
        &F_2:\; t_{ob}=\frac{\sum\left(t_b-t_a\right)}{t_F} \times 100 \%,
        \label{eq:t_ob}
        \end{align}
        where the subscripts $a$ and $b$ are the indices of timestamps satisfying $d_s \leq d_{\text {safe }}, a \leq k \leq b$, and $d_s$ representing the distance between any of the links and obstacles. 
For the validation use-case, we tailored $d_{\text{safe}}$ to be equal to the radius of the collision spheres $r_{cs}$. 
\newline
\textit{Smoothness} metrics quantify the oscillations in a robot's trajectory. For shared controllers, oscillations often result from the trade-off between reference tracking and obstacle avoidance. We identified three smoothness metrics, i.e., end-effector path smoothness, curvature change and velocity smoothness.
The end-effector path smoothness, $f_{p s}$,  quantifies oscillations of the end-effector in Cartesian space and its  expression is given by:
\begin{equation}
    F_3:\; f_{p s}= \frac{1}{S}\sum_{g=2}^{N-1}\left\|\boldsymbol{\Delta} \mathbf{p}_{g+1}-\boldsymbol{\Delta} \mathbf{p}_g\right\|^2,
    \label{eq:f_ps}
\end{equation}
where $g$ denotes the sample, $N$ the total number of samples, $\mathbf{p}_g$ is the end-effector Cartesian position at sample $g$ and S is the total travelled path computed as $S = \sum_{g=2}^{N} \left\|\boldsymbol{\Delta} \mathbf{p}_g\right\|$ which normalizes the path smoothness over the traveled distance. 
Curvature change is the second smoothness metric and it is complementary to path smoothness. Given the linear and angular Cartesian velocity, $v(t)$ and $\omega(t)$, of the end-effector the curvature equation becomes:
        \begin{equation}
            \kappa(t)=\left|\frac{\omega(t)}{v(t)}\right|.
        \end{equation}
        The curvature change metric as defined in \cite{calisi2009performance} is given by:
        \begin{equation}
            F_4:\; f_{cc}=\frac{1}{S}\int_0^{t_F}\left|\kappa^{\prime}(t)\right| d t,
            \label{eq:f_cc}
        \end{equation}
        where the measure is divided by the path length to normalize for different movements. 
\newline
The third smoothness metric measures velocity smoothness and captures oscillations at the joint level. This metric is relevant when the controller can influence multiple actuated joints. The metric, $f_{vs}$,  is calculated as the zero crossing rate of the acceleration for all joints composing the robot:
        \begin{align}
            & a_g =  \frac{v_{g+1}-v_g}{t_{g+1}-t_g}, \\
            F_5:\; & f_{v s} = 
            \frac{1}{N-1} \sum_{g=1}^{N-1} |sign(a_g)-sign(a_{g-1})|,
            \label{eq:f_vs}
        \end{align}
where $v_{g}$ denotes the joint linear velocity at sample $g$. The joint subscript is omitted. 
\newline
\textit{Efficiency} relates to the computation performance of the MPC controller and measures how fast, on average, the controller can compute a sample. The average computation time $t_{C}$ is defined as:
\begin{equation}
        F_6:\; t_{C} = \frac{1}{N} \sum_{g=1}^{N}t_{g+1}- t_{g},
        \label{eq:t_C}
\end{equation}
where $t$ is the time required to compute the sample, $g$ is the sample index and $N$ is the total number of samples. To summarize, Eq.~\eqref{eq:d_ob}, \eqref{eq:t_ob}, \eqref{eq:f_ps}, \eqref{eq:f_cc}, \eqref{eq:f_vs}, \eqref{eq:t_C} are the six performance metrics used in this work.
\subsection{User input representation}\label{sec:user_input}
\begin{figure}[b]
\centering
    \includegraphics[width=\linewidth]{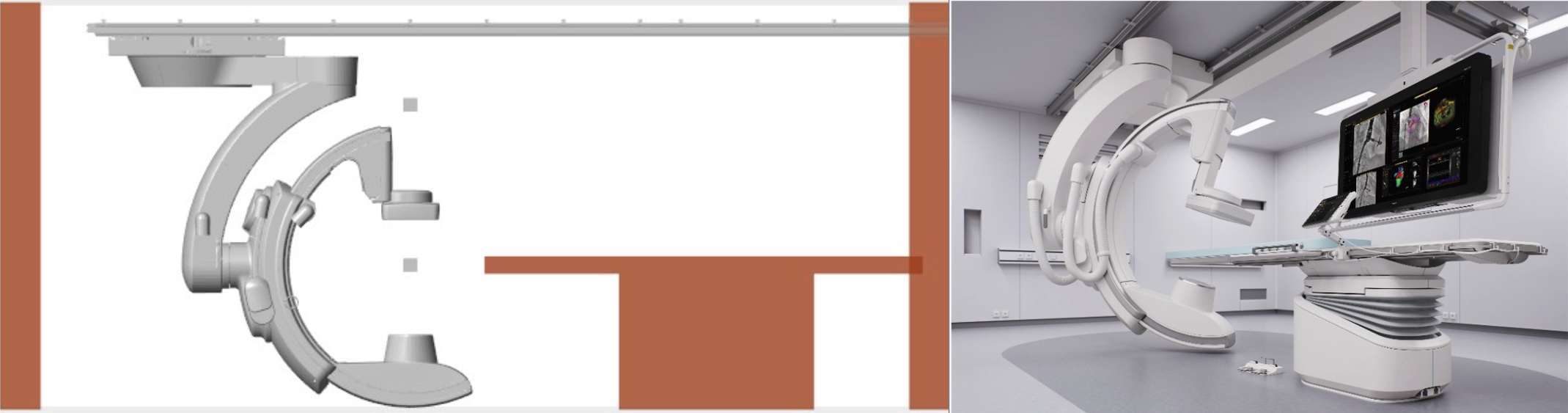}
\caption{The environment created for the simulations (left). An image of a real operating room (right).}
\label{fig:environment}
\end{figure}

\begin{figure}[t]
    \centering
    \includegraphics[ width =1\columnwidth]{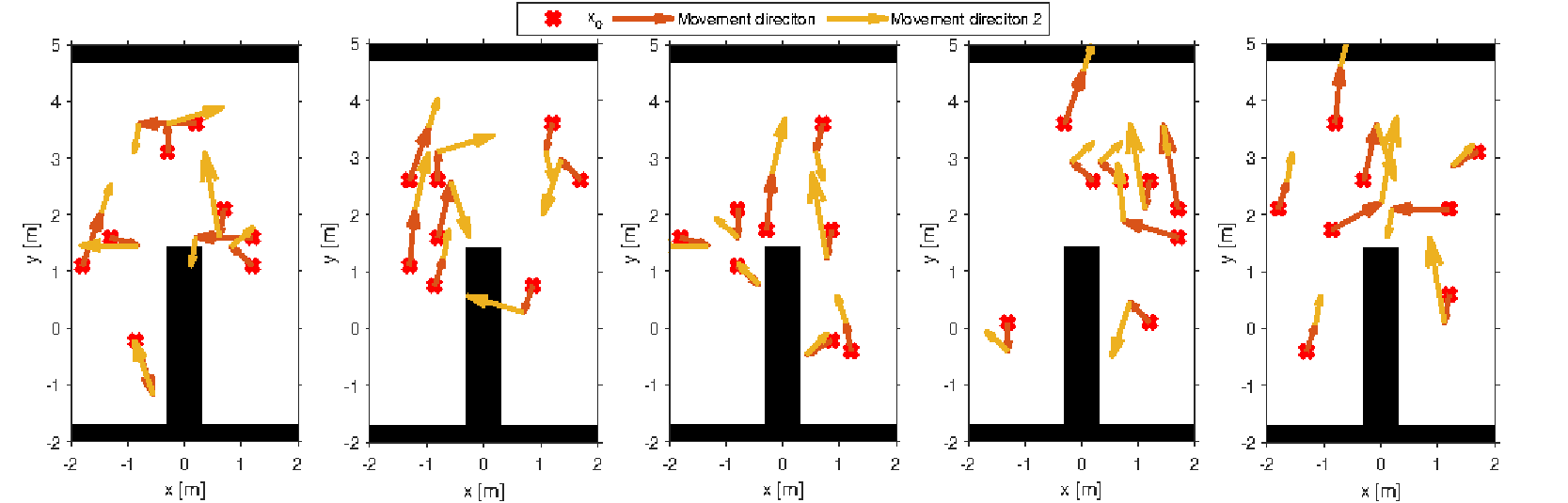}
    \caption{All concatenated velocity vectors for simulation. The first vector is in red, the second is in orange, and obstacles are in black.}
    \label{fig:movements}
    \vspace{-0.25in}
\end{figure}

The evaluation of the performance metrics requires gathering data about the performance of the robot during a movement under shared control. As we rely on simulation for data collection, our challenge lies in accurately representing user inputs to encompass a wide spectrum of scenarios, including both typical and corner cases.
For every movement, we propose to represent user inputs as a concatenation of $n$ velocity vectors which are applied each for a time $T/n$ where $T$ is the total time allowed to complete a movement.
The velocity vector can have different magnitudes and directions which is what is expected from typical inputs of joystick control. The concatenation of velocity vectors with different magnitudes and directions allows the simulation of different acceleration and deceleration profiles. The initial robot joints' configuration is assumed to be randomized for every movement which means that for some movements the application of the velocity input might result in a request to move towards obstacles. This is beneficial during optimization since it replicates edge cases that exemplify the tradeoff between following user commands and avoiding obstacles.
\newline
For the validation use-case, we concatenated two velocity vectors per movement and considered a maximum execution time $T$ of $20$s.
Three magnitudes were defined:  $\frac{k}{3}\boldsymbol{v}_{max}$ for $k = 1,2,3$, where $\boldsymbol{v}_{max}$ represents the maximum velocity magnitude the user can request via the joystick. 
A simulation environment was created to emulate rooms where IGT robots usually operate. The environment comprises a room with two walls and a patient table and was implemented in Simscape Multibody. The lateral view of the simulated room is visible in Fig~\ref{fig:environment}. In the same figure, an image of an actual room is also reported.
A total of $n_{mov}=40$ different concatenation options were generated. A topdown view of the room with superimposed all generated movements is displayed in Fig.~\ref{fig:movements}. 

\section{Experimental validation}\label{sec:validation} 
We applied the BO framework of Alg.~\ref{alg:BO}  to the automatic tuning of the MPC-based shared controller in Eq.~\ref{eq:stage_costs} for the IGT robot in Fig.\ref{fig:drawing_robot}. The performance of the optimized controller is compared to the performance of a baseline, hand-tuned, controller. Furthermore, we performed a VR-based user study to validate the generalization of the optimized controller to user input not provided during the automatic tuning procedure.
\newline
A weighted sum of the metrics is a crucial term in the definition of the objective function $J_{\xi}$ in Eq.~\ref{eq:BO}. The choice of the weights reflects the relative prioritization of the metrics and it is a task-dependent choice. For the validation use-case, their prioritization is reported in Table~\ref{tab:specs2}.
As mentioned in Section~\ref{sec:shared_control}, seven parameters were selected for the optimization procedure, i.e., $\boldsymbol{\xi}=[N_p, N_c, Q_{x}, Q_{y}, Q_{\theta}, c_1, c_2]$. Their values after the application of the BO procedure are reported in Tab.~\ref{tab:opt_var}. For comparison purposes, the values of those parameters for the baseline, hand-tuned controller, are reported as well.

 \begin{table}[b]
    \centering
    \caption{Overview of the metrics and their weights.}
    \label{tab:specs2}
    \begin{tabular}{|l l l c|} 
        \hline
        \textbf{Type} & \textbf{Metric} & $\boldsymbol{F_{hj}}$ & $\boldsymbol{w_{h}}$    \\ 
        \hline
        \multirow{2}{*}{Safety} & Obstacle proximity  & $d_{ob}$ [$\mathrm{m}$]  & 0.15	\\
        & Time spent near obstacles & $t_{ob}$ [$\%$] & 0.30\\
        \hline
        \multirow{3}{*}{Smoothness} & Path smoothness     &  $f_{ps}$ [$\mathrm{m}$]  & 	0.15\\  
        & Curvature change    & $f_{cc}$ [$\mathrm{rad/m}$]  &  0.25	\\
        &Velocity smoothness & $f_{vs}$ [$\mathrm{m/s^{2}}$] & 	0.10\\
        \hline
        Efficiency &  
        Computation time    & $t_C$ [$\mathrm{ms}$]     &  0.05	\\
        \hline
    \end{tabular}
\end{table}

\begin{table}[b]
    \centering
    \caption{The baseline and optimized MPC parameters.}
    \begin{tabular}{|l c c c c c c c|}
    \hline
         & $\boldsymbol{N_p}$ & $\boldsymbol{N_{c}}$ & $\boldsymbol{Q_{x}}$ & $\boldsymbol{Q_{y}}$ & $\boldsymbol{Q_{\theta}}$ & $\boldsymbol{c_{1}}$ & $\boldsymbol{c_{2}}$   \\ \hline
         base & 25 & 13 & 1 & 1 & 1 & 5 & 20  \\ \hline
         opt. & 21 & 5 & 3.047 & 0.62117 & 5.9981 & 7.3219 & 20.138 \\\hline
    \end{tabular}
    \label{tab:opt_var}
    \vspace{-0.25in}
\end{table}

A full optimization procedure could be executed on a computer with an Intel Core i7 10th generation processor running at 2.7 GHz in less than 120 minutes.
The performance metrics, defined in Section~\ref{sec:objective_function}, were evaluated for a set of 50 movements executed with the optimized MPC parameter set as well as with the hand-tuned set. The results are reported in Table~\ref{tab:results_simulation}.

\begin{table}[t]
    \centering
    \caption{Performance data for 50 simulated movements. The confidence interval is reported in brackets.}
    \label{tab:results_simulation}
    \small 
    \setlength{\tabcolsep}{4pt} 
    \renewcommand{\arraystretch}{1.2} 
    \begin{tabular}{|l c c c c c c |}
        \hline
         & $d_{ob}$  & $t_{ob}$    & $f_{ps}$  & $f_{cc}$  & $f_{vs}$ &   $t_C$\\ 
         & [$\mathrm{m}$] & [$\mathrm{\%}$]   &   [$\mathrm{m}$] &[$\mathrm{rad/m}$]  & [$\mathrm{m/s^{2}}$] &  [$\mathrm{ms}$]  \\ \hline
        base   & 0.68  & 0  & $1.2 \times 10^{-5}$ & 38.38 & 0.29 &  43.60\\ 
         & (0.05) & (0)  & ($1.6 \times 10^{-5}$) & (8.91) & (0.05) &  (1.82)  \\ \hline
        opt.  & 0.63 & 0 &  $1.1 \times 10^{-5}$ & 19.96 &  0.25 &  50.63 \\
         & (0.06) & (0) & ($1.4 \times 10^{-5}$) & (4.63) & (0.07) &   (2.26) \\\hline
    \end{tabular}
\end{table}
For every metric, the collected performance values were subject to a one-tailed t-test to assess if the observed difference between the optimized and the nominal parameter set is statistically significant. 
The optimized set returned an overall objective function improvement of 14\% with respect to the hand-tuned controller (from 0.42  to 0.36) without introducing any infeasible solution. Overall, the optimized set significantly improved the curvature change metric. No other metric was found to be statistically different. This result is in line with the weights associated to the curvature change metric (see Table~\ref{tab:opt_var}) which is the second largest after time spent near obstacles. 

\subsection{Comparison in Virtual Reality}
\begin{figure}[b]
    \centering
    \includegraphics[height = 0.55\columnwidth]{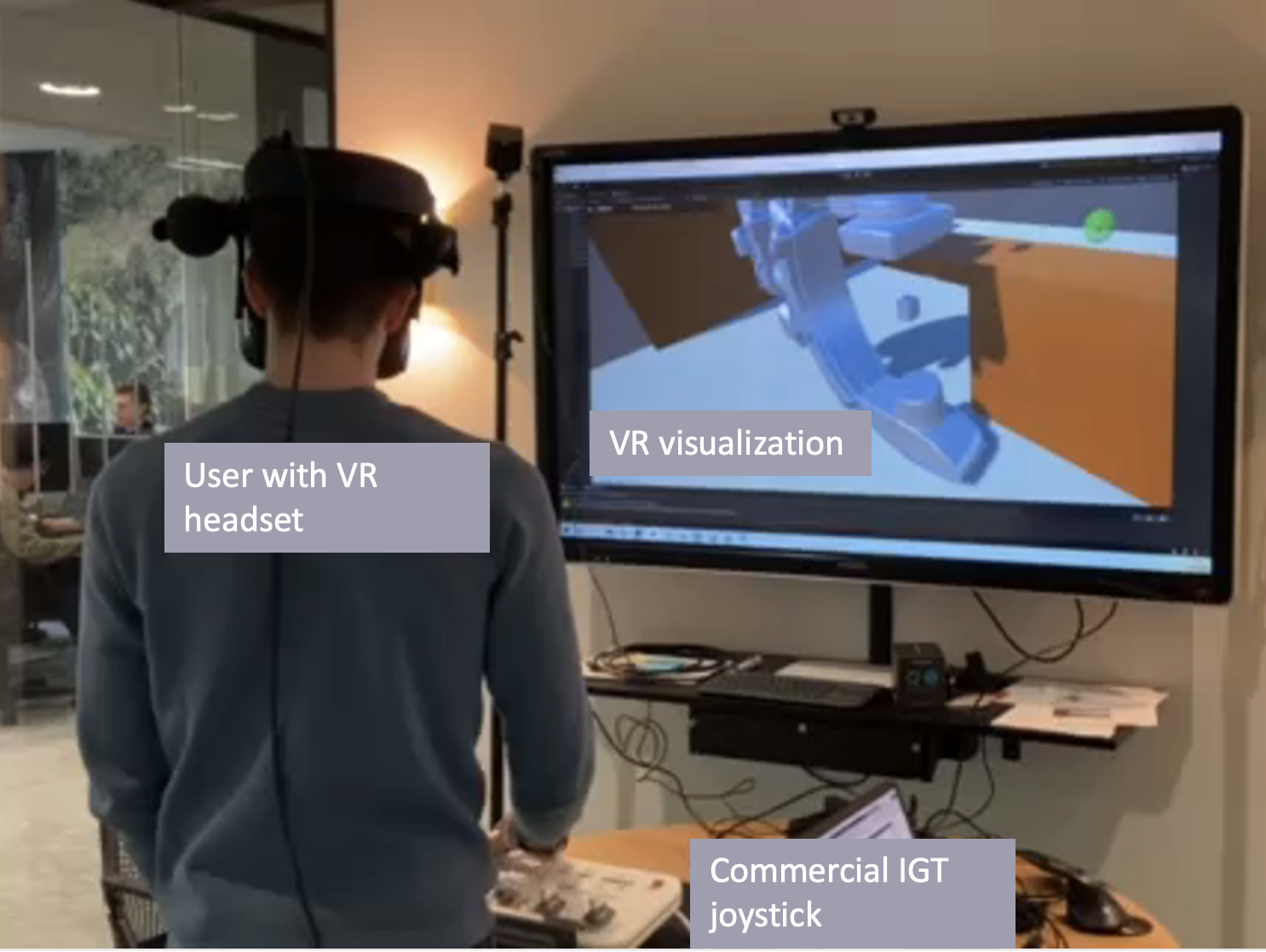}
    \caption{The setup of the VR  experiment.}
    \label{fig:vrsetup}
    \vspace{-0.25in}
\end{figure}

\begin{figure}[t]
    \centering
    \includegraphics[width = 1\columnwidth, trim=0 0 0 1.5cm, clip]{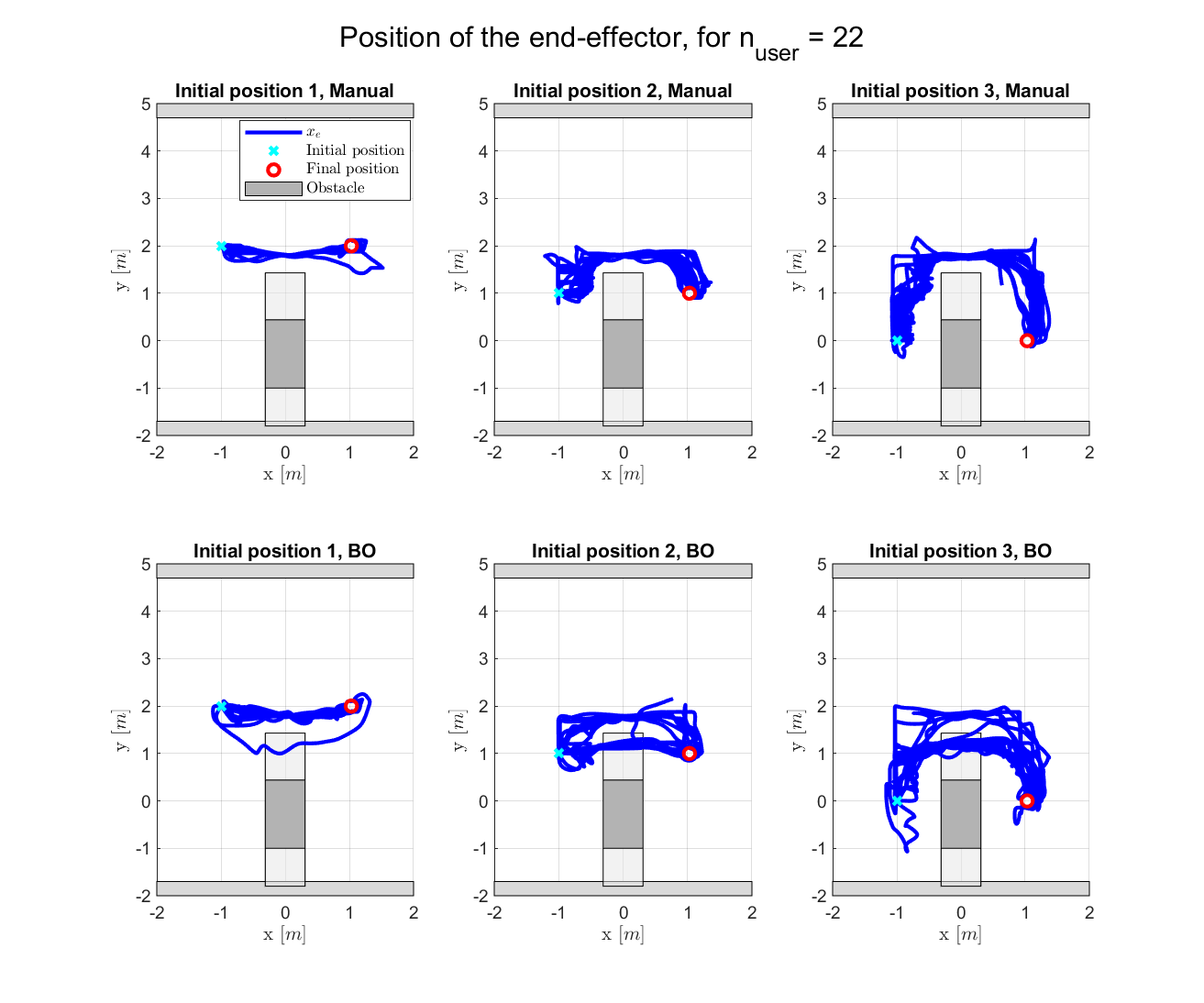}
    \caption{The end-effector path  for all users with the initial position (light blue circle) and the desired final position (red circle). Top hand-tuned controller. Bottom optimized controller.}
    \label{fig:pose_all}
    \vspace{-0.25in}
\end{figure}

To validate the generalization of the optimized controller for user inputs that were not provided during the optimization procedure, we tested the controller performance in a VR setup comparing the optimized controller with respect to the hand-tuned one.  
A VR simulation of the IGT robot was created which was connected to a  joystick-like control module.  The control module was used to gather user input in the form of desired velocity of the robot end-effector. Commands were communicated to  the MPC controller implemented in Matlab. The VR simulation was realized in Unity. A two way Matlab/Unity connection was established via a ROS bridge. The environment used for the simulations during optimization (see Fig.~\ref{fig:environment})  was  replicated exactly in Unity. During the experiment, users were requested to move the end-effector of the robot from an initial position on one side of the patient table to the opposite side, mimicking typical usage.  A visualization of the VR experimental setup is displayed in Fig.~\ref{fig:vrsetup}.
The experiment had a within-subject design. Two variables were manipulated during the experiment, the pair initial and final position of the end-effector (goal) and the controller configuration (hand-tuned or optimized). This led to a 3 (goals) x 2 (MPC configurations) experimental design which resulted in 6 trials per participant.
A total of 25 participants, with a mean age of 38 years, took part in the user study. However, data from 3 participants were excluded due to setup errors.
Approximately $73\%$ of the participants had prior experience with the IGT robot used in the study, while $45\%$ had prior experience with VR. The 6 trials were presented in random order. \newline
The resulting paths of the end-effector for all users for every experimental condition are displayed in Fig.~\ref{fig:pose_all}.
For every condition that configured the controller with the hand-tuned set, none of the users was able to move the end-effector on the other side of the table by letting the C-shaped link  traverse the tabletop (see Fig~\ref{fig:pose_all}, top graphs).  It was achievable, though, with the optimized set (see Fig~\ref{fig:pose_all}, bottom graphs).
Verbal user feedback provided during the experiment confirmed that the hand-tuned shared controller was too conservative w.r.t. repulsive actions in the vicinity of obstacles which was not the case for the optimized set.
By observing the paths obtained from the optimized set  (Fig~\ref{fig:pose_all}, bottom graphs) one can see that not every user achieved the goal position by traversing the tabletop. This is likely due to the design of the experiment itself, users that experienced the hand-tuned condition first were biased thinking that the robot could not traverse the tabletop and they did not try to do so in the follow-up trials. This was also confirmed by verbal feedback from users during the experiment.
\newline
The values of the metrics computed on the data set from the VR experiments are reported in Table~\ref{tab:results_vr}.
\begin{table}[b]
    \centering
    \caption{Performance data for the VR user experiments. The confidence interval is reported in brackets.}
    \label{tab:results_vr} 
    \small 
    \setlength{\tabcolsep}{4pt} 
    \renewcommand{\arraystretch}{1.2} 
    \begin{tabular}{|l c c c c c c |}
        \hline
          & $d_{ob}$  & $t_{ob}$   & $f_{ps}$ & $f_{cc}$  & $f_{vs}$ &  $t_C$\\ 
         & [$\mathrm{m}$] & [$\mathrm{\%}$]   & [$\mathrm{m}$] & [$\mathrm{rad/m}$] & [$\mathrm{m/s^{2}}$] &   [$\mathrm{ms}$]   \\ \hline
        base  & 0.55  & 0 &$2.4 \times 10^{-5}$ &  142.30 & 0.13  &   112.90 \\ 
         & (0.01) & (0) & ($3 \times 10^{-6}$) & (88.5) & (0.02) &   (0.64)  \\ \hline
        opt.  & 0.52 & 0.08 & $2.0 \times 10^{-5}$ & 49.11 & 0.22 &   105.30 \\
        & (0.02) & (0.16) & ($3.1 \times 10^{-6}$) & (30.05) & (0.02) &  (1.46)  \\\hline
    \end{tabular}
    \vspace{-0.25in}
\end{table}
Surprisingly, a one-tail t-test shows a statistically significant improvement in all metrics apart from time spent close to obstacles. We can argue that the difference in performance between simulation and VR is likely due to differences in the type of user input. Nevertheless, the optimized controller appears to have generalized to user-requested movements that were not provided during the optimization procedure.
A problem that emerged during VR validation is the number of infeasible solutions encountered when solving the MPC problem. Recall that no infeasible solutions were encountered for the simulated movements. In VR, the optimized controller encountered an infeasible solution on 9\% of its iterations, while the baseline controller only encountered them on 0.2\% of its iterations. During VR experiments, when infeasibility occurred, the robot kept executing the feasible effort computed in the previous sample and users autonomously adapted their input until a feasible solution was found again. Nevertheless, we can argue for the importance of reducing the number of infeasible solutions to the minimum. A possible improvement is the addition of real user inputs alongside simulated ones during optimization, at the likely cost of longer optimization time.  It is also important to notice that in one trial the distance between the robot and obstacles was less than the radius of the collision spheres. This is an unwanted situation and further underlines the need to improve user input representation during optimization.
\newline
The computation time $t_{C}$ of the optimized controller was reduced with respect to the baseline controller likely due to the shorter control horizon $N_{c}$.
\newline
For the smoothness metrics, the larger improvement is seen for the curvature change, as expected since that metric has the largest weight in its category.

\section{Conclusion}\label{sec:conclusion}
This paper presented a framework for the tailoring of Bayesian optimization to the automatic tuning of MPC-based shared controllers, including a proposal for performance metrics and representation of user inputs. The performance of the tuned controller has been validated in simulation and with a VR-based user study. Results showed a significant performance improvement of the optimized controller w.r.t. a hand-tuned version. While tuning in simulation allowed to rapidly reach an optimized controller configuration, the user study outlined further challenges for the generalization of the results. Future work will look at enhancing the representation of the user input during the optimization process, possibly complementing it with data from experiments with users. While the proposed methods effectively reduced the performance metrics, it can be argued that other factors might be important in shared control such as user-perceived comfort or ergonomics. Extending the work considering an enlarged set of metrics constitutes another interesting research direction. 





\bibliographystyle{IEEEtran}
\bibliography{bibliography}

\end{document}